%% file: main_arxiv.tex
\documentclass[10pt,twocolumn,letterpaper]{article}

\usepackage[pagenumbers]{cvpr}

\usepackage{enumitem}
\usepackage{epigraph}
\usepackage{amsmath}
\usepackage{xspace}
\usepackage{algorithm}
\usepackage[noend]{algpseudocode}
\usepackage{booktabs}
\usepackage{tabularx}
\usepackage{makecell}
\usepackage[normalem]{ulem}
\usepackage{mdframed}
\usepackage{listings}
\usepackage{xcolor}

\usepackage{amssymb}
\usepackage{graphicx}

\usepackage[pagebackref,breaklinks,colorlinks]{hyperref}

\newcommand{\citet}{\cite}

\usepackage[capitalize]{cleveref}
\crefname{section}{Sec.}{Secs.}
\Crefname{section}{Section}{Sections}
\Crefname{table}{Table}{Tables}
\crefname{table}{Tab.}{Tabs.}

\def\cvprPaperID{*****} 
\def\confName{CVPR}
\def\confYear{2022}

\input{macro}

\begin{document}

\title{Zero-shot Sequential Neuro-symbolic Reasoning for Automatically Generating Architecture Schematic Designs}

\author{Milin Kodnongbua*\\
University of Washington\\
{\tt\small milink@cs.washington.edu}
\and
Lawrence H. Curtis\\
WinnDevelopment\\
{\tt\small lcurtis@winnco.com}
\and
Adriana Schulz\\
University of Washington\\
{\tt\small adriana@cs.washington.edu}
}


\twocolumn[{%
\renewcommand\twocolumn[1][]{#1}%
\maketitle
\begin{center}
    \centering
    \captionsetup{type=figure}
      \includegraphics[width=\textwidth]{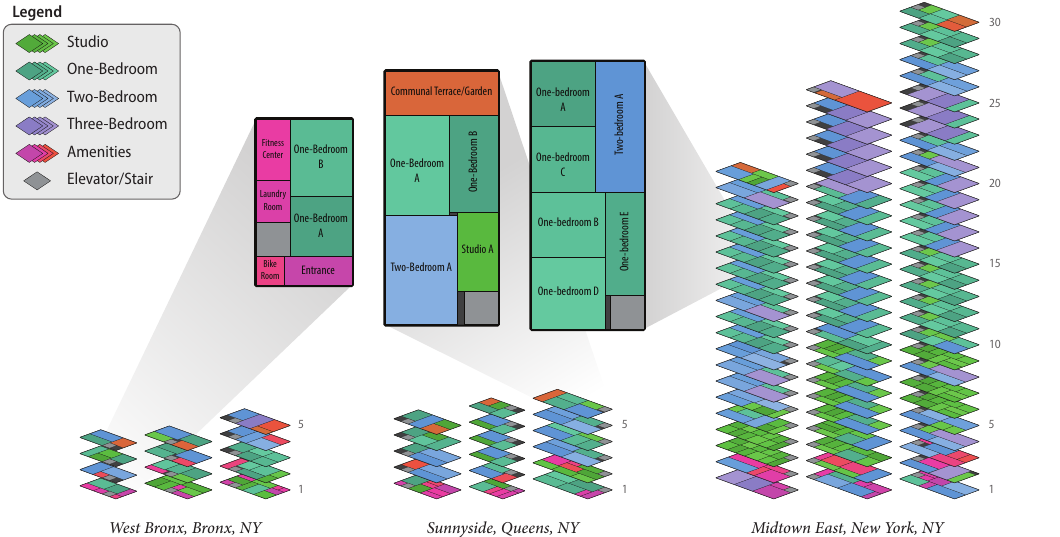}
      \caption{Our automatically generated apartment floor plans for three different locations: Bronx, Queens, and Manhattan, NY. Three different runs are shown for each location. The given lot dimension is 40 by 100 feet.}
    \label{fig:teaser}
\end{center}%
}]

\input{abstract}

\input{sec/1_introduction}
\input{sec/2_related_work}
\input{sec/3_methods}
\input{sec/4_results}

\input{sec/5_limitations}
\input{sec/6_conclusion}


{\small
\bibliographystyle{ieee_fullname}
\bibliography{bibliography,schulz,cad,arch,textto3d,llm,floorplangen}
}

\newpage
\input{sec/7_figures}

\end{document}

%% file: macro.tex
\newif\ifshowauthorcomments

\showauthorcommentsfalse
\newcommand{\authorcomment}[3]{\ifshowauthorcomments{\color{#3}~\{#1: #2\}}\fi}

\newcommand{\milin}[1]{\authorcomment{MK}{#1}{cyan}}

\newcommand{\stkout}[1]{\ifmmode\text{\sout{\ensuremath{#1}}}\else\sout{#1}\fi}

\newif\ifseechanges

\ifseechanges

\else

\fi

%% file: abstract.tex
\begin{abstract}
This paper introduces a novel automated system for generating architecture schematic designs aimed at streamlining complex decision-making at the multifamily real estate development project's outset. Leveraging the combined strengths of generative AI (neuro reasoning) and mathematical program solvers (symbolic reasoning), the method addresses both the reliance on expert insights and technical challenges in architectural schematic design. To address the large-scale and interconnected nature of design decisions needed for designing a whole building, we proposed a novel \textit{sequential neuro-symbolic reasoning approach}, emulating traditional architecture design processes from initial concept to detailed layout. To remove the need to hand-craft a cost function to approximate the desired objectives, we propose a solution that uses neuro reasoning to generate constraints and cost functions that the symbolic solvers can use to solve. We also incorporate feedback loops for each design stage to ensure a tight integration between neuro and symbolic reasoning. Developed using GPT-4 without further training, our method's effectiveness is validated through comparative studies with real-world buildings. Our method can generate various building designs in accordance with the understanding of the neighborhood, showcasing its potential to transform the realm of architectural schematic design.

\end{abstract}

%% file: sec/1_introduction.tex
\section{Introduction}





The initial phase in multifamily real estate development encompasses many high-level decisions, such as the identification of construction requirements, potential sites, costs, and more. An integral part of this phase involves crafting a schematic design for a building; that is, given an understanding of a potential building site, coming up with a rough draft of what the building might look like. Such a schematic is important for paving the way for deeper analysis, and allows real estate developers and property owners to ultimately make decisions about what to build and where. This paper endeavors to assist developers by automating the creation of such schematic designs, allowing multiple solutions for multiple sites to be generated quickly and hence supporting high-level decision making.


Building upon the complexities involved in crafting a schematic design for a building, which requires architects to balance their expertise in building amenities, apartment types, dimensions, and resource utilization with the constraints of structural requirements, we introduce a method that marries two distinct yet complementary approaches. This method combines the innovative prowess of generative AI, adept at distilling expert insights and referred to as \textit{neuro reasoning}, with the structured rigor of formal verification, ensuring that designs adhere to necessary constraints, known as \textit{symbolic reasoning}. This holistic approach aims to streamline the intricate process of architectural design, addressing both the creative and technical aspects seamlessly.

In architectural schematic design, a primary challenge is the complexity and interdependency of the decisions involved. These decisions encompass a wide range of factors, such as the amenity choices and the distributions of units of different sizes over each floor.  Although state-of-the-art foundation models have made significant strides in encapsulating much of the architectural expertise, they struggle with scaling to address all these decisions simultaneously for large-scale buildings. Moreover, each decision requires not only creative insight but also verification of multiple requirements. Therefore, in contrast to simpler design tasks that might be divided into two distinct phases – one leveraging neuro reasoning and the other symbolic reasoning, potentially linked by a feedback loop~\cite{makatura:2023} – the field of architectural schematic design necessitates a method adept at managing both the extensive variety and the intricate interconnectivity of these decisions.

To address this challenge, we propose a solution inspired by traditional design principles of sequential decision-making, an approach commonly used by architects in schematic design. It involves breaking the problem into a series of decisions, from high-level to low-level, each informing the subsequent stage of design. 
While such a breakdown in optimization-driven generative design would lead to a multi-level program with nested optimization loops, our integration of neuro-reasoning circumvents this by prioritizing inference over optimization. This approach more effectively mirrors the natural, sequential process of design. We refer to this proposed approach as \emph{sequential neuro-symbolic reasoning}.

Our method is based on two fundamental observations. Firstly, we recognize that the challenge in decision making using purely symbolic solutions lies in the need for \emph{appropriate cost functions}. In most computational design problems, such cost functions are hand-crafted heuristics that are based on expertise in the domain in an attempt to approximate true objectives. In some cases, the objective can be difficult to computationally describe. For example, how could one define a cost function to describe how suitable a set of amenities is for an apartment in Los Angeles? To address this, we integrate neuro-reasoning in decision making, allowing us to shift the focus of our symbolic reasoning engine from heuristic-driven optimization to structural satisfaction over objectives and constraints defined by the AI model itself.


Secondly, since we integrate solvers in our system in intermediate stages, extra care needs to be taken to ensure our inference is valid. We note that by dividing the process into neural and symbolic components, we indirectly introduce cost models, which are problematic for sequential reasoning. Our insight is to bridge the information gap between the neuro and symbolic approaches by establishing feedback loops that connect the output of the symbolic solver back to the neuro-reasoning steps. These loops ensure that our decisions remain consistent with the objectives set by the AI model.

We build our tool using GPT-4 \cite{openai2023gpt4} without further training and a commercial mathematical program solver, Gurobi \cite{gurobi}. To validate our method, we evaluate our solution by contrasting it against alternative approaches using ablation studies and juxtapose our outcomes with those from real buildings. Our method outputs a variety of buildings while maintaining the key aspects that define the neighborhood. In most cases, the generated buildings match well with those in real life. The findings demonstrate the generative capability and usefulness of our approach.

%% file: sec/2_related_work.tex
\section{Related Work}


\paragraph{Generative AI in Computational Designs}


Recent research has investigated the potential of harnessing pre-trained large language models (PLMs) to reduce reliance on training data across various domains, such as robotics \cite{mirchandani2023large}, embedded systems \cite{englhardt2023exploring}, and manufacturing-oriented design \cite{makatura:2023}. These studies identify the inherent challenges in symbolic reasoning within PLMs, underscoring the need for innovative integration of symbolic solvers. Our work concentrates on architectural schematic designs and introduces a novel method combining sequential reasoning with neuro-symbolic approaches. 

\paragraph{Symbolic Reasoning with PLMs}
In recent years, PLMs have gained significant interest in the research community due to their impressive capability in an array of tasks. However, mathematical and symbolic reasoning still remains one of the most significant challenges of PLMs \cite{patel2021nlp}. Several prompting strategies have been proposed to improve the performance in this regard: few-shot prompting (prompting with a few input-output exemplars without fine-tuning) \cite{brown2020llmfewshot, chen2021evaluating, chowdhery2022palm}; and strategies asking PLMs to generate their reasoning steps, such as chain-of-thought \cite{wei2023chainofthought}, scratchpads \cite{nye2021work}, and least-to-most \cite{zhou2023leasttomost}. More recently, \citet{chen2023program} and \citet{gao2023pal} require PLMs to output reasoning steps as a Python code instead, allowing them to leverage the rigor of the Python interpreter to output accurate answers. \citet{parisi2022talm} and \citet{schick2023toolformer} propose a strategy to use any text-based external tools, such as web APIs or symbolic functions, by asking PLMs to identify appropriate function and its arguments based on the problem descriptions and parse the function's output into natural language answers. In fact, ChatGPT already has this functionality built-in and can call various external tools, such as Wolfram, web search, and Python functions. Our work is similar to this regime. However, our work not only uses a symbolic solver in a sequential manner to tackle complex design problems but also uses feedback loops to bridge the information gap between the neuro and symbolic approaches. 


\paragraph{PLMs in Architecture}

While there is a body of research in the integration of AI techniques with construction and architecture industry, the integration of PLMs to enable natural language interactions with architectural systems remains underexplored. Several works have explored the use of PLMs to retrieve information from a BIM model using natural language \cite{wu2019bimretrieval, elghaish2022bimvoice, zheng2023bimgpt}. \citet{jang2023interactive} uses ChatGPT to make design changes to a BIM model by converting BIM to XML format and asking ChatGPT to modify the XML. \citet{Prieto_2023} uses ChatGPT to generate a construction schedule for a simple construction project. We refer the reader to the survey in the use of GPT models in the construction industry by \citet{saka2023gpt}. The closest to our work is Architext which generates a floor plan of a apartment unit from textual description, e.g. ``an apartment with three bedrooms and two bathrooms'' \cite{galanos2023architext}. It fine-tunes a GPT model to output coordinates and dimensions of rooms to compose the floor plan. Our method addresses a different problem: taking a high-level description of the lot to generate floor plans of the entire apartment building. Ours also differs in that we can guarantee the validity of the floor plans (e.g., rooms are non-intersecting), and we do not further fine-tune GPT-4 to do the task. 






\paragraph{Automated Generation of Floor Plans}
Automated floor plan generation is a well explored area of research. A floor plan creation method takes a series of geometric constraints such as area, adjacency, and boundary and outputs a floor plan that is optionally optimized for some objectives such as daylight, gap spaces, or heating cost. The method can be divided into three big categories. Bottom-up methods arrange smaller components (e.g. rooms) into larger structures (e.g. floors). \cite{merrel2010bdglayoutgen, bahrehmand2017layoutopt, bisht2022graphfloorplan}. Top-down methods subdivide the building boundary into smaller units \cite{michalek2002layoutopt,marson2010treemap,rodrigues2014floorplangen}. Data-driven methods utilize machine learning models trained on real floor plan datasets \cite{wu2019datafloorplan,hu2020graph2plan,nauata2021housegan}. We refer to \citet{weber2022floorplangen} for a complete literature review of automated floor plan generation algorithms.
We note that these methods relies on detailed objectives and constraints, requiring expertise in the field. Our novel approach can work with high-level site specifications to generate floor plans.
While our implemented solver simplifies the design space (not considering non-rectangular lots and units), future work could expand this scope by integrating symbolic solvers from listed literature into our novel system.

%% file: sec/3_methods.tex
\begin{figure*}[t!]
    \centering
    \includegraphics[width=\linewidth]{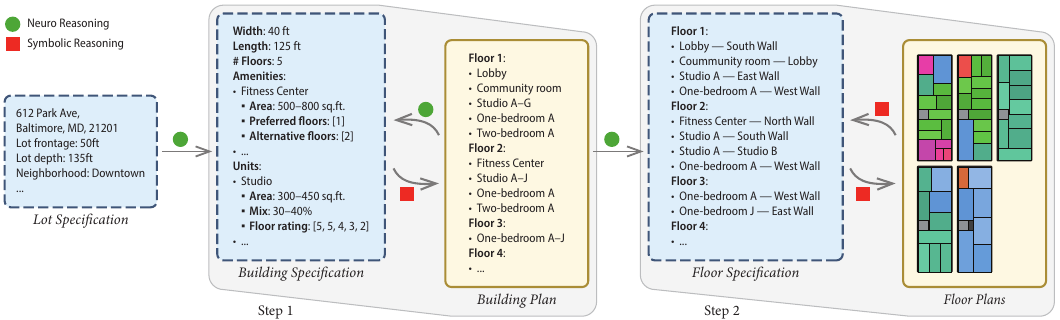}
    \caption{Overview of our method. Given high-level lot specification, it outputs building floor plans. Our method is a two-step process. In Step 1, we generate a \textit{building plan} consisting of amenities and units to include on each floor, and in Step 2, we generate the building \textit{floor plans}. In each step, we use neuro reasoning to create \textit{specifications} which can then be solved using symbolic reasoning. Our method is further enhanced by feedback loops that are based on neuro reasoning in Step 1 and symbolic reasoning in Step 2.}
    \label{fig:overview}
\end{figure*}

\section{Methods}


Our system accepts a textual description of a lot, which includes its dimensions, location, and, optionally, information about the neighborhood. It outputs building floor plans that comprise the building's footprint (width, length, and number of floors) and the layout of each floor. This layout is divided into \textit{elements} that can be either private (apartment units of different sizes) or communal (such as a lobby, fitness center, or other amenities). 


As previously discussed, the fundamental challenges in automatically generating such detailed plans are the involvement of multiple and interdependent considerations 
and the difficulty in formalizing them as objectives.
Our \textit{sequential neuro-symbolic} address these challenges with two key insights.


Firstly, our approach employs sequential reasoning, a method inspired by the hierarchical decision-making process typical in architectural design. We observe that while it is challenging to define overarching objectives for intermediate stages, \emph{constraints can be checked and imposed at each step}. For example, we can validate the total area usage before doing the layout of a floor. Doing so reduces the complexity of the problem. We therefore propose a two-step solution, as illustrated in Figure~\ref{fig:overview}. In Step 1, we decide the building footprint and the \textit{elements} to be included on each floor, which we call a \textit{building plan}; and in Step 2, we decide the placement of these \textit{elements}, outputting the building \textit{floor plans}.

Secondly, our approach uses a neural approach to make decisions that would otherwise require architectural expertise and are hard to specify symbolically and uses a symbolic approach for ones that requires computations and spatial reasoning. We propose a neuro-symbolic approach where in each step, we have the PLM generate a \textit{specification}, a list of constraints and objectives, and have a symbolic solver finds a solution that conforms to the \textit{specification}.


However, the PLM, without being fully capable of predicting the output of the solvers, might not output \textit{specifications} that optimizes the final outcomes. We propose to add feedback loops that allows the PLM to adjust its \textit{specifications} after it sees the solver's output. Our insight is that such loops help the PLM implicitly learn the behavior of the solvers and make informed decisions.
In Step 1, we do a neural feedback loop where the PLM gives feedback on the \textit{building plan}. In Step 2, a more concerning problem is the infeasibility of the \textit{specification}. We propose a symbolic feedback loop where the solver provides several conflicting parts of the \textit{specification}, from which the PLM can choose to remove to avoid infeasibility.

\subsection{Building Plan Generation}


Recall that a \textit{building plan} consists of the building footprint and a list of \textit{elements} on each floor. Different neighborhoods may have different zoning classes and regulations which limit the size and number of floors of the building. They may have different demographics which demand for specific unit types and amenities. For example, in a town with many families, several industrial plants, and no colleges, it is preferable to include more two- and three-bedrooms and fewer one-bedrooms. Financial consideration can also be a factor in determining the unit mix. For example, having more studios and one-bedrooms in New York City can be more profitable. \milin{We could ask Larry for more insights} These considerations require extensive research on the neighborhood and can be handled by the PLM. On the other hand, constraints such as the total area occupied by the \textit{elements} on each floor must not exceed the available floor area can be handled using a solver.

\subsubsection{Building specification}
A \textit{building specification} collectively describes information the PLM needs to output. It consists of the number of floors, width and length of the building, a list of \textit{amenity specifications}, and a list of \textit{unit specifications}. An \textit{amenity specification} contains the type of the amenity, the area range it could occupy, a list of preferred floors, and a list of alternative floors. A \textit{unit specification} contains the type of the unit (e.g., studio, one-bedroom, etc.), the percentage range of total unit count, area range, and the likelihood (from 0 to 5) of putting the unit on each floor. 

\subsubsection{Solving for a Building Plan}
The goal for the solver is to determine the number of units for each type such that the ratio is within the specified range and assign each unit and amenity to one of the floors while ensuring that we have enough space and have a wide area range for each floor, which gives the flexibility when solving for the floor plans. We then optimize for amenities being on the preferred floors and units being on floors with a high likelihood score. 
We formulate this as a mixed integer linear programming (MILP) and solve using Gurobi. If the solver could not find a solution, we ask the PLM to regenerate the \textit{building specification}, potentially widening the area and unit ratio ranges. In addition, we perform a preliminary check to see immediate infeasibility when generating floor plans, which we also regenerate the specification if it fails. 

\subsubsection{Neural feedback loop}

The neural feedback loop allows the PLM to modify the \textit{building specification} once it sees the \textit{building plan}. Specifically, we ask the PLM if there are concerns about the balance of the amenities and units and allow it to suggest modifications on the list of \textit{amenity specifications}. It can remove existing amenities, modify the area or floor preference of existing amenities, or propose a new amenity. We then run the solver again and repeat until the PLM does not suggest any further modifications.

\subsection{Floor Plan Generation}
In this step, we decide the location and size of each amenity and unit on each floor. Such decisions require understanding of their functionality. Amenities that are likely used together are better if they are placed closed together. For example, it makes sense to have a mail room and bike storage close to the entrance to minimize walking. Outdoor spaces should be at the boundary of the building to access sunlight and outdoor scenery. These adjacency considerations can be generated by the PLM. On the other hand, constraints like no two \textit{elements} should not overlap can be handled by the solver.

\subsubsection{Floor Specification}
A \textit{floor specification} contains adjacency constrains on each floor. It describes whether two \textit{elements} should be adjacent or an \textit{element} should be on any boundary (north, east, south, west) of the building.

\subsubsection{Solving for Floor Plans}
We model every \textit{element} as an axis aligned rectangle where the design variables are its position and dimension. In addition, we allocate an elevator/stair space that spans vertically across all floors, and make the position of the elevator/stair another design variable. In addition to adjacency constraints in \textit{floor specification}, we constrain that no two \textit{elements} intersect, the ratio between width and height does not exceed 2.7, and a unit has to be adjacent to one of the walls to accommodate at least one window. We maximize the space utilization. We formulate this problem as a mixed integer quadratic programming (MIQP) problem and solve using Gurobi. We solve each floor individually, and use the position of the elevator/stair on the first floor on every other floor.

\subsubsection{Symbolic Feedback Loop}
The symbolic feedback loop points out a set of conflicting constraints in the \textit{floor specification} and asks the PLM to remove one of them. The process repeats until the floor plans are generated. To achieve this, we ask Gurobi to output an irreducible inconsistent subsystem (IIS), the smallest subset of constraints that make the problem infeasible. If the IIS contains conflicting adjacency constraints, we can precisely give a list of conflicting constraints. However, if the infeasibility is due to different reasons, we fallback to a full regeneration of the \textit{floor specification}.

\subsection{Interfacing with GPT-4}
To extract information from GPT-4, we ask it to output a JSON string which can easily be parsed. We provide an example JSON with the desired format and populate the fields with generic values. For example, to get the \textit{amenity specification}, we ask:
\begin{verbatim}
Output in the following JSON format:
```json
[{"type": "a", "min_area": 1, "max_area": 1,
  "preferred_floors": [1],
  "alternative_floors": [2, 3]}]
```
\end{verbatim}
When parsing responses, we perform thorough checks for the data types and value ranges and provide precise feedback on any parsing error. We repeat the correction process for up to five times.

We refer the readers to our supplementary material for further implementation details, symbolic problem formulations, and prompts.

%% file: sec/4_results.tex
\begin{figure*}[t!]
    \centering
    \includegraphics[width=\linewidth]{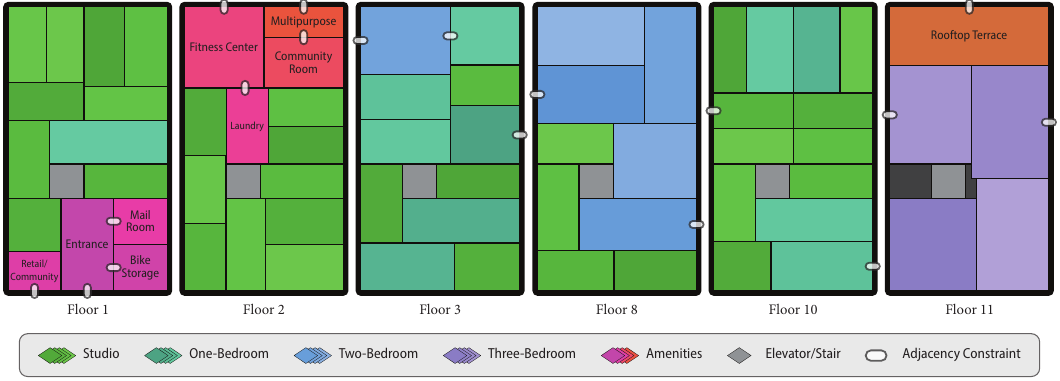}
    \caption{Some of the generated floor plans for 360 W 43rd St, New York, NY (1). The generated adjacency constraints group the amenities together and put the rooftop terrace at the boundary of the building. The generated plans also exhibit high utilization of the available space.}
    \label{fig:floorplan}
\end{figure*}

\section{Results}\label{sec:results}

We evaluate our system by its ability to make informed decisions (building footprint, amenity choices, and unit mixes) with respect to different locations. We conduct two case studies: one on hypothetical lots in New York city and another on five actual apartment buildings across the US. In addition, we run multiple ablation studies to formalize the need for each contribution in our \emph{sequential neuro-symbolic reasoning} approach.

We use GPT-4 and Gurobi 10.0.2 for our \emph{neuro} and \emph{symbolic reasoning} engines. All experiments were conducted on a machine with a 48-core CPU. Each generation took about 5--60~minutes depending on the number of floors and feedback iterations. Please see the supplementary material for the full transcripts with GPT-4 and generated floor plans for each run.

\subsection{Case Study 1: Hypothetical Lots in New York City}
In this study, we run our system on the same lot dimension of 40 by 100 feet but for different locations. We choose three locations: West Bronx, Bronx; Sunnyside, Queens; and Midtown East, Manhattan. 

The resulting floor plans are illustrated in Figure~\ref{fig:teaser}. The generated building footprint ranges from 4--5~floors in the Bronx, 5--6~floors in Queens, to 20--30~floors in Manhattan. These match the zoning regulations of R5-R7 in Bronx and Queens and R10 in Manhattan. In Table~\ref{tab:hyp-result}, we list the unit mixes and amenities of each building. We also include actual unit distributions in each area, processed from the US Census Bureau data, as a reference. 
It is important to note that these census figures represent an aggregate and not the unit distribution of individual buildings, making them a broad reference rather than a direct comparison. In Manhattan, where there are more units per building, our unit mix closely matches the census data. In contrast, in the Bronx and Queens, our findings differ more, showing a higher prevalence of studios and one-bedrooms and fewer three-bedrooms. However, the small number of units in these buildings means that even adding a few three-bedroom units would markedly alter the percentages, indicating a still reasonable alignment with overall trends.
In terms of amenities, most buildings include basic amenities such as fitness center, bike storage, rooftop terrace, laundry room, and community room. However, buildings in Manhattan tend to include more amenities such as a business center, a children's playroom, and a pet facility. Overall, we can see consistent trends in the building footprints, amenity selections, and unit mixes across different locations.

\subsection{Case Study 2: Comparisons with Actual Buildings}
We now want to compare our generated buildings against real world examples. We selected five actual apartments in Downtown, Baltimore, MD; Capitol Hill, Seattle, WA; Lakeview, Chicago, IL; Brentwood, Los Angeles, CA; and Midtown West, New York, NY. We will refer to these buildings and specific neighborhoods by their city for the rest of the discussion. Due to the limitations of our floor plan generation algorithm, we had to limit our search to buildings that are rectangular in shape and have a lot width of less than 60 feet. The lot dimensions were measured using Google Maps. We also added short paragraphs for demographics and commute options of the neighborhood to the lot description. For comparison, we curated the list of amenities, the number of floors, and the number of units from sources available online. However, information regarding the unit mixes is generally not available.

We summarize various attributes of the generated and real buildings in Table~\ref{tab:real-result}. 
In most cases, the number of floors is a close match 
In the case of Chicago, the neighborhood contains multiple zoning classes. Most areas allow for only mid-rise buildings except for a narrow strip of land near Lake Michigan which our building happens to be in that allows for taller buildings.
For unit mixes, it is worth noting that the buildings in the Seattle lot have a balanced unit mix whereas buildings in the Los Angeles lot generally have more two- and three-bedrooms and fewer studios, which are consistent with the census data. In Baltimore and New York, we see more studios whereas in Chicago, we see evenly distributed mixes. Although the unit mixes are not always mirroring the census data, they do reflect clear, location-specific trends.
In terms of amenities, our system typically includes more amenities than those of actual buildings, and most amenities in real buildings can be found on most generated buildings. We observe that GPT-4 likes to include certain amenities to all buildings: fitness center, bike storage, rooftop terrace, community room, and package room. However, it makes specialized decisions like including co-working spaces in all Baltimore and Seattle buildings and pet facilities in Seattle buildings. 

In most cases, we suspect that the inaccuracies are attributed to GPT-4 not having enough information. 
We believe that our system can be more accurate if GPT-4 is further trained with more data.
Overall, our system can output variations of number of floors, unit mixes, and amenities across different runs in each location, highlighting the generative capability of our method.


\subsection{Floor Plan Quality}
We discuss the GPT-4 generated \textit{room specifications} and the solver generated \textit{floor plans}. We show the generated floor plans of one building in New York, NY in Figure~\ref{fig:floorplan}. In the input \textit{lot specification}, we specified that the building entrance has to be on the south end of the building. GPT-4 correctly outputs this as the constraint for 17 out of 24 runs. We suspect the remaining cases were due to the long distance between the input and when the \textit{room specification} is generated, and GPT-4 might have paid less attention to that information. In most cases, GPT-4 added adjacency constraints among amenities like the package room, bike storage, community room, and lobby, making several cliques of amenities. In terms of the \textit{floor plans}, our solver performs well in maximizing space utilization achieving 98.2\% on average. While we acknowledge there are other architectural constraints we have not taken into account, we demonstrate the ability of our system to generate reasonable floor plans, which can be used as inspirations for architects to generate building designs.





\subsection{Ablation 1: Sequentiality}
We justify the need to break the problem down to multiple stages by comparing it to an approach that directly asks GPT-4 to output the locations for each \textit{element} in one single step. In the prompt, we put a step-by-step instruction on intermediate decisions to make (e.g., number of floors, building dimension, unit mix, etc.) and constraints to impose (e.g., \textit{elements} must be non-overlapping, elevators must be at the same position on all floors, etc.). We also include the output format of the floor plans, which is a list of \textit{element} types, positions, and dimensions, in the prompt. Figure~\ref{fig:ablation}a shows the generated floor plans of three different runs with the lot size of 40 by 100 feet in Manhattan. GPT-4 outputs reasonable choices for the number of floors (20). It sometimes did basic arithmetic to compute total \textit{element} area versus the available floor area. However, it usually skipped the calculation and assigned random combinations of \textit{elements} in each floor causing poor utilization of the floor space. We also notice that GPT-4 will only make computations on the first few items saying the rest can follow the pattern. For example, it only outputs the first three floor plans. Nonetheless, GPT-4 performed decently in arranging the \textit{elements} on the floor. It can put one \textit{element} to the right or bottom of existing \textit{elements}. However, generating floor plans is a complicated problem; enforcing that all \textit{elements} do not intersect and are within the boundary seems to exceed the capability of GPT-4, which is expected of a language model.


\subsection{Ablation 2: Symbolic Reasoning}
In this experiment, we still purely use GPT-4, but we break down the problem as described in the method section. We first generate the \textit{building specification}, then \textit{building plan}, and finally the \textit{floor plans}. However, instead of using a solver in computational tasks, we list out constraints that need to be met and ask GPT-4 to validate its own outputs and propose new solutions if they fail. For the floor plans, we just ask GPT-4 in the same manner as in Ablation 1 without validation since it would be too difficult for GPT-4 to validate 2D plans. Figure~\ref{fig:ablation}b shows the results of this method again on a hypothetical lot in Manhattan. Visually, we see more sophisticated floor plans than those in Ablation 1. We notice a better space utilization and unit selection which is due to the separate step to validate the \textit{building plan}. However, these self validation steps have several pitfalls. GPT-4 does not always make valid arithmetic comparisons. For example, given that the total number of studios has to be within 10--15, it outputs ``Total Units Count Check: Studio: 2nd Floor (3) + 3rd Floor (3) = 6 (Within the range 10-15)'', which is clearly incorrect. This is also the reason we see \textit{elements} occupying significantly more or less space than the floor allows in some runs. GPT-4 also likes to perform validations to only the first few items which makes it not feasible for large-scale problems. Again, GPT-4 made good attempts in generating floor plans, but we still see overlapping \textit{elements} and elevators being out of place. Overall, the two ablation experiments suggest the need for a sequential approach and the need for a symbolic reasoning engine to handle various constraints however simple they are.

\subsection{Ablation 3: Neural Feedback Loop}
We now discuss the performance of the neural feedback loop that looks at the \textit{building plan} and adjusts the \textit{building specifications} with regard to the amenity choices. While comparing two sets of amenities one before and after the adjustments can be subjective, we highlight some interesting behaviors. Majority of the feedback was about adjusting the sizes of amenities to match the building size and occupancy. In large buildings like in Manhattan, GPT-4 suggested enlarging community spaces like the fitness center and outdoor space to match the high unit count whereas in smaller buildings like in Seattle, we see suggestions to downsize those spaces. We also notice adjustments happening over multiple iterations. For example, in one building in Los Angeles, the area of the rooftop terrace was first downsized from 1,200 to 700 sq.ft. and further reduced to 600 sq.ft.. In some iterations, we observe suggestions to add more niche amenities like pet spa and children's playroom if the space allows, and we often see area reduction of other amenities in subsequent iteration to maintain overall ratio between amenities and units. Interestingly, we see suggestions to include a pet facility in all three buildings in Seattle citing that the Capitol Hill neighborhood is pet-friendly. We also notice multiple cases of combining two amenities like community room and co-working space into one multipurpose room to improve space usage efficiency and multiple cases of removing laundry room in favor of in-unit washers and dryers. Altogether, the neural feedback loop is essential in that it allows GPT-4 to re-adjust the specification after it sees a concrete number of units. This feedback loop can also be viewed as the optimization of the \textit{building plan} towards what GPT-4 thinks is the best with respect to the neighborhood and building footprint. 





\subsection{Ablation 4: Symbolic Feedback Loop}
We justify the need for our symbolic feedback loop that points out specific adjacency constraints that causes infeasibility when generating the floor plans. Out of 24 runs in both case studies, 13 runs had infeasible initial adjacency constraints, and our feedback loop can correct 10 of those runs without regenerating the entire set of constraints. We include additional results of our method without both feedback loops in the supplemental material. While regenerating the \textit{floor specifications} eventually leads to feasible sets of constraints, it disregards results from previous iterations no matter how close they are to success. GPT-4 also has the tendency to reduce the number of constraints as the regeneration goes on. Instead, our feedback loop directly pinpoints conflicting constraints and GPT-4 has to remove one constraint. This approach preserves GPT-4's original intention and can be viewed as more sample efficient.

%% file: sec/5_limitations.tex
\section{Limitations and Future Work}
\label{sec:limitations}


\paragraph{Support for Architectural Considerations}
When designing and constructing residential multi-story buildings, there are numerous considerations and objectives that we currently do not cover in this work. For example, it is more efficient to line up bathrooms across multiple floors to minimize the hassle for pipes and ventilation equipment; and in larger lots, we can have different non-rectangular building massings. Recent advances in floor plans generation tools already have better support for such considerations \cite{weber2022floorplangen}. However, these methods take a specific set of constraints and objectives and do not take high-level lot specifications like ours does. 
Our fundamental idea would still apply as future work adapts these tools to interface with our system. This would enhance the quality of the results and make our method more versatile.


\paragraph{GPT-4 as the Prior}
Although GPT-4 have general idea about neighborhoods, it does not know specific information as we pinpoint the address, which are crucial in determining several attributes of the buildings. Future work could further train GPT-4 on more specific information. One could create a streamlined architectural knowledge base and use Retrieval Augmented Generation (RAG) method \cite{lewis2021retrievalaugmented} to query relevant and accurate information. Querying information given specific geolocation is another interesting research direction. Such advancements would help our system to generate more accurate buildings.

\paragraph{Vision Feedback Loop}
While it would be ideal to have the PLM give feedback on the generated floor plans and improve upon itself, this is difficult because floor plans are 2D graphics which cannot be interpreted by the PLM. 
Future work could incorporate a vision model to parse the floor plans and make suggestions about the room placement on the floor.

\paragraph{Interfacing with PLMs}
While PLMs can generate JSON strings, it remains a challenge to control the schema and guarantee the values are valid. 
We found few instances of GPT-4 breaking the interface. For example, GPT-4 once put ``B1'' as the preferred floor for bike storage. In the neural feedback loop, GPT-4 suggested changes to the unit mix while we only allow modifications to the amenities; it added a new amenity type named ``One Bedroom'' which we could not validate and correct. While this limitation of PLMs is an interesting problem of its own, we acknowledge that such errors were not unreasonable and were partly due to our choice of the interface. Future work could extend the interfaces to support underground floors and modification to unit mixes among other things.

\paragraph{Interactions with Experts}
An interesting aspect of our proposed tools is that it can generate a large number of different plans that a developer can consider. In the future, it would be helpful to allow users to give feedback and iterate on the design process. For example, users can say what they like from each building and the system would generate a new building following the preferences. Further improvement could allow the model to focus on more diversity when considering a variety of possible solutions. Another direction would be to have users in the loop using GPT-4 as the interface to the symbolic solver. For example, users can add additional adjacency constraints or move certain amenities to certain floors using a natural-language text description.


%% file: sec/6_conclusion.tex
\section{Conclusion}
In this paper, we present a novel sequential neuro-symbolic approach to tackle the complex task of generating apartment floor plans which can be useful assets for architects.
While a traditional apartment design process requires expertise in the area, our approach breaks new ground by leveraging foundational models to formalize relevant considerations and symbolic techniques to actualize them into structurally valid designs. Our results demonstrate our method's generative capability and its potential to transform real estate development workflows. 

Our work also lays a foundation for future research in generative design. Traditional methods grapple with challenges such as high computational costs and the complexity of defining clear, measurable design goals. In contrast, recent advancements in machine learning approach design as an inference process, sidestepping these issues but at the expense of constraint imposition and performance objective maximization. Our proposed neuro-symbolic system showcases the potential of combining these methodologies and opens up exciting avenues for future exploration.


%% file: sec/7_figures.tex
\clearpage

\begin{table*}[t!]
\caption{Attributes of the generated buildings in Bronx, Queens, and Manhattan, NY on a hypothetical lot of size 40 by 100 feet. Unit distributions in the area are included for comparison.}
\footnotesize
\begin{tabular}{r|rrrr|rrrr|rrrr}
\toprule
& \multicolumn{4}{c|}{West Bronx, Bronx} & \multicolumn{4}{c|}{Sunnyside, Queens} & \multicolumn{4}{c}{Midtown East, New York} \\
\cmidrule{2-13}
& \multicolumn{1}{c}{Real} & \multicolumn{1}{c}{1} & \multicolumn{1}{c}{2} & \multicolumn{1}{c|}{3} & \multicolumn{1}{c}{Real} & \multicolumn{1}{c}{1} & \multicolumn{1}{c}{2} & \multicolumn{1}{c|}{3} &\multicolumn{1}{c}{Real} & \multicolumn{1}{c}{1} & \multicolumn{1}{c}{2} & \multicolumn{1}{c}{3} \\
\midrule
Width &  & 34 & 40 & 40 &  & 40 & 36 & 40 &  & 40 & 40 & 40 \\
Length &  & 59 & 70 & 75 &  & 70 & 55 & 90 &  & 95 & 100 & 90 \\
\cmidrule(lr){1-13}
\# Floors &  & 4 & 4 & 5 &  & 5 & 6 & 6 &  & 20 & 25 & 30 \\
\# Units &  & 10 & 15 & 22 &  & 14 & 16 & 26 &  & 111 & 120 & 145 \\
\cmidrule(lr){1-13}
Studio &  & 2 & 7 & 7 &  & 3 & 5 & 7 &  & 32 & 36 & 43 \\
1B &  & 5 & 6 & 10 &  & 6 & 7 & 11 &  & 45 & 48 & 58 \\
2B &  & 3 & 2 & 4 &  & 5 & 4 & 8 &  & 28 & 24 & 29 \\
3B+ &  & 0 & 0 & 1 &  & 0 & 0 & 0 &  & 6 & 12 & 15 \\
\cmidrule(lr){1-13}
Studio (\%) & 11\textsuperscript{\textdagger} & 20 & 47 & 32 & 17\textsuperscript{\textdagger} & 21 & 31 & 27 & 23\textsuperscript{\textdagger} & 29 & 30 & 30 \\
1B (\%) & 36\textsuperscript{\textdagger} & 50 & 40 & 45 & 43\textsuperscript{\textdagger} & 43 & 44 & 42 & 45\textsuperscript{\textdagger} & 41 & 40 & 40 \\
2B (\%) & 36\textsuperscript{\textdagger} & 30 & 13 & 18 & 28\textsuperscript{\textdagger} & 36 & 25 & 31 & 24\textsuperscript{\textdagger} & 25 & 20 & 20 \\
3B+ (\%) & 17\textsuperscript{\textdagger} & 0 & 0 & 5 & 12\textsuperscript{\textdagger} & 0 & 0 & 0 & 9\textsuperscript{\textdagger} & 5 & 10 & 10 \\
\cmidrule(lr){1-13}
Fitness Center &  & \checkmark & \checkmark & \checkmark &  & \checkmark & \checkmark & \checkmark &  & \checkmark & \checkmark & \checkmark \\
Bike Storage &  & \checkmark & \checkmark & \checkmark &  & \checkmark & \checkmark & \checkmark &  & \checkmark & \checkmark & \checkmark \\
Rooftop Terrace &  & \checkmark & \checkmark & \checkmark &  & \checkmark & \checkmark & \checkmark &  & \checkmark & \checkmark & \checkmark \\
Laundry &  & \checkmark & \checkmark & \checkmark &  & \checkmark & \checkmark & \checkmark &  &  &  & \checkmark \\
Community Room &  & \checkmark &  & \checkmark &  & \checkmark & \checkmark & \checkmark &  &  & \checkmark & \checkmark \\
Package Room &  &  &  & \checkmark &  & \checkmark &  &  &  &  & \checkmark & \checkmark \\
Storage &  &  &  &  &  & \checkmark &  &  &  &  & \checkmark &  \\
Multipurpose Room &  &  &  &  &  &  &  &  &  & \checkmark &  & \checkmark \\
Children Playroom &  &  &  &  &  &  &  &  &  & \checkmark &  & \checkmark \\
Pet Facility &  &  &  &  &  &  &  &  &  &  & \checkmark & \checkmark \\
Business Center &  &  &  &  &  &  &  &  &  & \checkmark &  &  \\
\bottomrule
\multicolumn{13}{p{10cm}}{\textdagger \; Unit distribution by the number of bedrooms in the area. Source: 2022 American Community Survey Public Use Microdata Sample, US Census Bureau. (Note: numbers are not averaged by buildings)}
\end{tabular}
\label{tab:hyp-result}
\end{table*}

\begin{table*}[t!]
\footnotesize
\caption{Attributes of the generated buildings compared to five real buildings.}
\resizebox{\linewidth}{!}{%
\begin{tabular}{r|rrrr|rrrr|rrrr|rrrr|rrrr}
\toprule
 & \multicolumn{4}{c|}{611 Park Ave} & \multicolumn{4}{c|}{1815 Bellevue Ave} & \multicolumn{4}{c|}{350 W Oakdale Ave} & \multicolumn{4}{c|}{11649 Mayfield Ave} & \multicolumn{4}{c}{360 W 43rd St}\\
 & \multicolumn{4}{c|}{Baltimore, MD} & \multicolumn{4}{c|}{Seattle, WA} & \multicolumn{4}{c|}{Chicaco, IL} & \multicolumn{4}{c|}{Los Angeles, CA} & \multicolumn{4}{c}{New York, NY}\\
 & \multicolumn{4}{c|}{(Downtown)} & \multicolumn{4}{c|}{(Capitol Hill)} & \multicolumn{4}{c|}{(Lakeview)} & \multicolumn{4}{c|}{(Brentwood)} & \multicolumn{4}{c}{(Midtown West)}\\
 \cmidrule{2-21}
 & \multicolumn{1}{c}{Real} & \multicolumn{1}{c}{1} & \multicolumn{1}{c}{2} & \multicolumn{1}{c|}{3} & \multicolumn{1}{c}{Real} & \multicolumn{1}{c}{1} & \multicolumn{1}{c}{2} & \multicolumn{1}{c|}{3} &\multicolumn{1}{c}{Real} & \multicolumn{1}{c}{1} & \multicolumn{1}{c}{2} & \multicolumn{1}{c|}{3} &\multicolumn{1}{c}{Real} & \multicolumn{1}{c}{1} & \multicolumn{1}{c}{2} & \multicolumn{1}{c|}{3} &\multicolumn{1}{c}{Real} & \multicolumn{1}{c}{1} & \multicolumn{1}{c}{2} & \multicolumn{1}{c}{3}\\
 \midrule
 \# Floors & 12 & 7 & 10 & 10 & 6 & 6 & 5 & 6 & 14 & 5 & 5 & 6 & 3 & 3 & 3 & 4 & 11 & 11 & 12 & 35 \\
\# Units & 110 & 60 & 100 & 70 & 65 & 35 & 30 & 44 & 148 & 33 & 27 & 42 & 11 & 14 & 11 & 11 & 172 & 105 & 68 & 328 \\
\cmidrule(lr){1-21}
Studio & N/A & 24 & 33 & 34 & 65 & 10 & 8 & 17 & N/A & 10 & 7 & 12 & N/A & 2 &  &  & N/A & 51 & 26 & 131 \\
1B & N/A & 24 & 47 & 29 &  & 14 & 13 & 18 & N/A & 13 & 11 & 17 & N/A & 6 & 4 & 5 & N/A & 32 & 28 & 164 \\
2B &  & 12 & 20 & 7 &  & 11 & 6 & 9 &  & 10 & 9 & 13 &  & 6 & 5 & 4 & N/A & 16 & 14 & 33 \\
3B+ &  &  &  &  &  &  & 3 &  &  &  &  &  &  &  & 2 & 2 & N/A & 6 &  &  \\
\cmidrule(lr){1-21}
Studio (\%) & 13\textsuperscript{\textdagger} & 40 & 33 & 49 & 30\textsuperscript{\textdagger} & 29 & 27 & 39 & 16\textsuperscript{\textdagger} & 30 & 26 & 29 & 5\textsuperscript{\textdagger} & 14 &  &  & 23\textsuperscript{\textdagger} & 49 & 38 & 40 \\
1B (\%) & 54\textsuperscript{\textdagger} & 40 & 47 & 41 & 44\textsuperscript{\textdagger} & 40 & 43 & 41 & 37\textsuperscript{\textdagger} & 39 & 41 & 40 & 26\textsuperscript{\textdagger} & 43 & 36 & 45 & 48\textsuperscript{\textdagger} & 30 & 41 & 50 \\
2B (\%)  & 28\textsuperscript{\textdagger} & 20 & 20 & 10 & 24\textsuperscript{\textdagger} & 31 & 20 & 20 & 30\textsuperscript{\textdagger} & 30 & 33 & 31 & 52\textsuperscript{\textdagger} & 43 & 45 & 36 & 21\textsuperscript{\textdagger} & 15 & 21 & 10 \\
3B+ (\%) & 5\textsuperscript{\textdagger} &  &  &  & 3\textsuperscript{\textdagger} &  & 10 &  & 17\textsuperscript{\textdagger} &  &  &  & 17\textsuperscript{\textdagger} &  & 18 & 18 & 7\textsuperscript{\textdagger} & 6 &  &  \\
\cmidrule(lr){1-21}
Fitness Center & \checkmark & \checkmark & \checkmark & \checkmark &  & \checkmark & \checkmark & \checkmark & \checkmark & \checkmark & \checkmark & \checkmark & \checkmark & \checkmark & \checkmark & \checkmark & \checkmark & \checkmark & \checkmark & \checkmark \\
Bike Storage &  & \checkmark & \checkmark & \checkmark & \checkmark & \checkmark & \checkmark & \checkmark & \checkmark & \checkmark & \checkmark & \checkmark & \checkmark &  & \checkmark & \checkmark & \checkmark & \checkmark & \checkmark & \checkmark \\
Rooftop Terrace &  &  & \checkmark & \checkmark & \checkmark & \checkmark & \checkmark & \checkmark &  & \checkmark & \checkmark & \checkmark & \checkmark & \checkmark &  & \checkmark & \checkmark & \checkmark & \checkmark & \checkmark \\
Community Room & \checkmark & \checkmark & \checkmark & \checkmark & \checkmark & \checkmark & \checkmark & \checkmark &  &  & \checkmark & \checkmark & \checkmark & \checkmark & \checkmark & \checkmark &  & \checkmark &  & \checkmark \\
Package Room &  & \checkmark & \checkmark & \checkmark & \checkmark &  & \checkmark & \checkmark &  & \checkmark & \checkmark & \checkmark & \checkmark &  & \checkmark & \checkmark & \checkmark & \checkmark &  & \checkmark \\
Laundry Room & \checkmark & \checkmark & \checkmark & \checkmark &  & \checkmark &  &  & \checkmark &  &  &  &  &  &  &  & \checkmark & \checkmark &  & \checkmark \\
Co-working Space &  & \checkmark & \checkmark & \checkmark &  & \checkmark & \checkmark & \checkmark &  & \checkmark &  &  &  &  & \checkmark &  &  &  & \checkmark &  \\
Pet Facility &  &  &  &  &  & \checkmark & \checkmark & \checkmark &  &  &  & \checkmark &  &  & \checkmark &  &  &  & \checkmark & \checkmark \\
Multipurpose Room &  &  & \checkmark &  &  &  &  &  &  &  &  &  &  & \checkmark &  &  & \checkmark & \checkmark & \checkmark &  \\
Outdoor Space &  & \checkmark &  &  &  &  &  &  &  &  &  &  &  &  &  &  & \checkmark &  &  & \checkmark \\
Storage &  &  &  &  &  &  &  &  & \checkmark &  &  &  & \checkmark &  &  &  &  &  &  &  \\
Pool & \checkmark &  &  &  &  &  &  &  &  &  &  &  &  &  &  &  &  &  &  &  \\
Shared Kitchens &  &  &  &  & \checkmark &  &  &  &  &  &  &  &  &  &  &  &  &  &  &  \\
Spa &  &  &  &  &  &  &  &  &  &  &  &  & \checkmark &  &  &  &  &  &  &  \\
Meeting Room &  &  &  &  &  &  &  &  &  &  &  &  &  & \checkmark &  &  &  &  &  &  \\
Retail &  &  &  &  &  &  &  &  &  &  &  &  &  &  &  &  &  & \checkmark &  &  \\
Recycling Center &  &  &  &  &  &  &  &  &  &  &  &  &  &  &  &  &  &  &  & \checkmark \\
Entertainment Room &  &  &  &  &  &  &  &  &  &  &  &  &  &  &  &  &  &  &  & \checkmark \\
 \bottomrule
 \multicolumn{21}{p{15cm}}{\textdagger\; See footnote of Table~\ref{tab:hyp-result}.}
\end{tabular}
}
\label{tab:real-result}
\end{table*}

\begin{figure*}[t!]
    \centering
    \includegraphics[width=\linewidth]{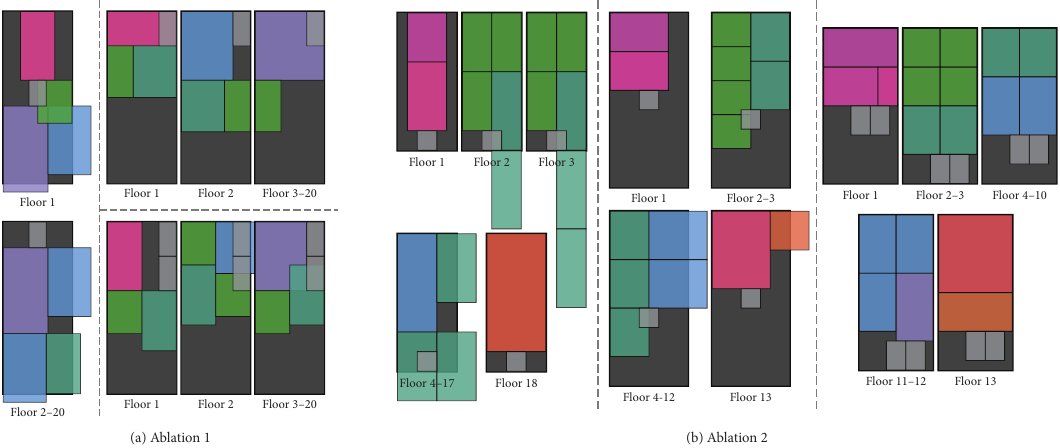}
    \caption{Generated floor plans using (a) only GPT-4 in a single step (Ablation 1); and (b) only GPT-4 breaking the problem to multiple steps (Ablation 2).}
    \label{fig:ablation}
\end{figure*}

\begin{figure*}[t!]
    \centering
    \includegraphics[width=\linewidth]{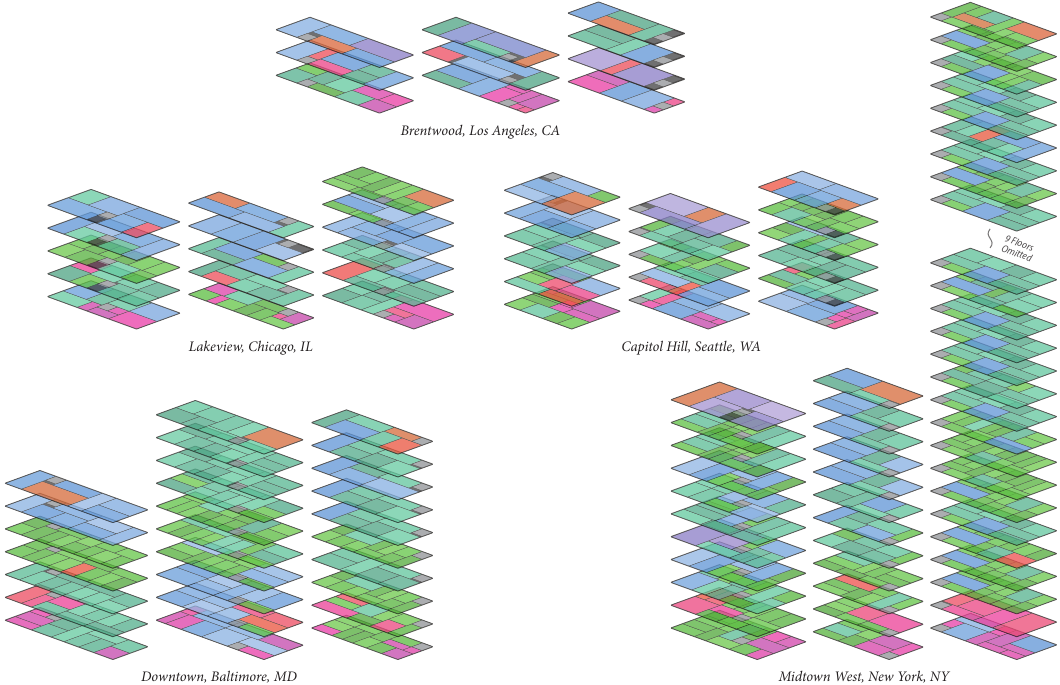}
    \caption{Gallery of generated apartment floor plans in five different neighborhoods: Brentwood, Los Angeles, CA; Lakeview, Chicago, IL, Capitol Hill, Seattle, WA; Downtown, Baltimore, MD; and Midtown West, New York, NY. The result shows variations in the building designs and trends for each location.}
    \label{fig:results}
\end{figure*}



